# Behavior Trees vs Executable Ontologies: A Comparative Analysis of Robot Control Paradigms

Alexander Boldachev

boldachev@gmail.com

https://orcid.org/0000-0002-7259-2952

## Abstract

This paper compares two distinct approaches to modeling robotic behavior: imperative Behavior Trees (BTs) and declarative Executable Ontologies (EO), implemented through the boldsea framework. BTs structure behavior hierarchically using control-flow, whereas EO represents the domain as a temporal, event-based semantic graph driven by dataflow rules. We demonstrate that EO achieves comparable reactivity and modularity to BTs through a fundamentally different architecture: replacing polling-based tick execution with event-driven state propagation. We propose that EO offers an alternative framework, moving from procedural programming to semantic domain modeling, to address the semantic-process gap in traditional robotic control. EO supports runtime model modification, full temporal traceability, and a unified representation of data, logic, and interface - features that are difficult or sometimes impossible to achieve with BTs, although BTs excel in established, predictable scenarios. The comparison is grounded in a practical mobile manipulation task. This comparison highlights the respective operational strengths of each approach in dynamic, evolving robotic systems.

## 1. Introduction

Modern robotic systems demand flexible, reactive, and easily modifiable control architectures. Behavior Trees (BTs), an imperative framework, have emerged as the industry standard for modeling complex autonomous behaviors, leveraging their intuitive hierarchical structure and modularity in both gaming and robotics applications.

Robotic environments are becoming more complex, demanding dynamic prioritization, adaptation to unforeseen events, and semantic transparency. This highlights limitations in control-flow architectures, which can lead to rigid and hard-to-maintain structures, especially for context-dependent behavior or integration with diverse agents. As an alternative, Executable Ontologies, implemented via the boldsea framework, offer a declarative approach based on event semantics (Boldachev, 2025)[1].

In earlier work on executable ontologies (Boldachev, 2025), the term semantic–process gap was introduced to denote the structural disconnection between semantic knowledge representations (maps, ontologies, task models) and the imperative control architectures that actually drive robot behavior. In that context, the gap was analyzed for BPM and RDF/OWL systems; here we show that an analogous gap appears in robotics when Behavior Trees are layered on top of separate world models. Executable Ontologies (EO) are designed to close this gap by unifying domain modeling and process execution in a single temporal semantic graph executed by an event-based dataflow engine (Boldachev, 2025). In contrast, BTs provide robust, hierarchical control suited for many applications.

The goal of this work is to conduct a systematic comparison of BTs and EO architectures using a practical mobile robot scenario. We evaluate both approaches against formal metrics derived from prior Behavior Trees and Fault-Tolerant Finite State Machines comparisons (Iovino et al., 2024), focusing on modularity, reactivity, and readability. Our primary contribution is to demonstrate, through working implementations, that the same task can be solved by two different methods: procedural control-flow (BTs) versus declarative dataflow (EO). We further analyze specific use cases where each paradigm excels, providing a clear framework for technology selection.

**Contributions.** First, through working implementations, we show that the same robotic control task can be solved by fundamentally different architectures: algorithmically (BTs) and as declarative dataflow models (EO). Second, we show that metrics originally developed for control-flow architectures (BTs vs. FSMs) have limited applicability when one of the compared systems uses a dataflow paradigm.

The remainder of the paper is organized as follows. Section 2 reviews the two architectures; Section 3 presents the benchmark task and its implementations; Section 4 compares them; Section 5 discusses implications; Section 6 concludes. The Appendix provides detailed examples of EO code and screenshots of the interfaces.

## 2. Background

To enable an objective comparison between Behavior Trees (BTs) and Executable Ontologies (EO), we examine the theoretical foundations and core operational principles of each architecture. Understanding their fundamental differences - namely, the imperative nature of BTs versus the declarative nature of EO - is essential for evaluating their respective strengths and weaknesses in practical robotics applications.

---

[1] In this paper we use the term Executable Ontologies (EO) in two related senses. At the conceptual level, EO denotes a general paradigm in which semantic models are directly executable. At the implementation level, all examples are instantiated using the boldsea framework as a concrete event-driven realization of this paradigm.

## 2.1. Behavior Trees: A Hierarchical Control-Flow Architecture

Behavior Trees (BTs) are a hierarchical, directed-tree structure designed to manage task execution and switching. BTs provide an intuitive visual representation of control logic, which has led to their widespread adoption as an industry standard for implementing complex autonomous behaviors.

The architecture is built from several key node types that define execution logic. Control nodes - such as *Sequence* and *Fallback* - orchestrate the flow of execution, particularly task switching. Execution nodes form the leaves of the tree, implementing concrete operations (*Actions*) or evaluating environmental states (*Conditions*).

BT execution is *tick-based*: at a fixed frequency, a signal (the "tick") is sent to the root node and propagates downward through the tree according to the logic of the control nodes. This continuous top-down, left-to-right traversal ensures high system reactivity. Consequently, BTs embody a procedural paradigm, modeling robot behavior as a sequence of steps, checks, and alternative actions.

## 2.2. Executable Ontologies: A Dataflow Architecture Based on Event Semantics

Executable Ontologies (EO) - as realized in the boldsea framework - offer a declarative and semantically rich approach to robot control.. In EO, every change in the system is recorded as an event within a directed, acyclic temporal graph. Each event is uniquely generated according to a predefined model, which semantically defines its type, permissible values, and activation conditions. This ensures runtime validation and a naturally structured temporal history. Execution follows a dataflow principle: when a new event is added to the graph, the system automatically evaluates dependent conditions (e.g., `Condition`, `SetValue`) without a central polling mechanism. Control logic is expressed declaratively through restrictions attached to model events: Condition defines when an event can occur, SetValue computes and assigns a value based on the current graph state, and `SetDo` triggers system actions (e.g., creating or modifying individuals). Thus, rather than programming a sequence of actions, the developer models the domain as a semantic graph; robot behavior emerges naturally as the consequence of these declarative rules responding to changes in the world state. EO leverages unified vocabularies to ensure semantic interoperability across systems.

# 3. Problem Statement for Comparative Analysis

## 3.1. Scenario: Delivery, Recharging, and Docking

To analyze the described approaches to modeling robot behavior, we employ the scenario proposed by Iovino et al. (2024) for the practical comparison of Behavior Trees (BTs) and Fault-Tolerant Finite State Machines (FSMs). The approach developed in that paper enables the evaluation of key architectural aspects: sequential action execution, handling of external asynchronous events, and dynamic priority switching. The scenario comprises a base task and two extensions:

**Base Task: Delivery**. The robot must perform a simple logistics operation: navigate to the object's location (`objectLoc`), grasp the object (`took`), transport it to the target location (`targetLoc`), place the object (`put`), and mark task completion (`delivered`).

**Extension 1**. **Recharging Task**. At any point during execution of the base task, a critical situation may arise: the robot's battery level (`batteryLevel`) drops below a threshold value (`batteryMin`). In this case, the robot must interrupt the delivery task and initiate recharging: navigate to the charging station location (`station`) and perform battery charging (`charging`). Upon completion of charging, the robot must resume the interrupted delivery task.

**Extension 2: Docking Task**. After successfully delivering the object (`delivered`), the robot must return to its docking station (`Loc Dock`).

This scenario is a comprehensive testbed, as it requires the system not only to follow a predefined plan but also to react flexibly to internal state changes (`battery level`) and to extend its behavior through model augmentation.

We begin the comparative analysis by examining the Behavior Trees implementation.

## 3.2. Behavior Trees Implementation

The detailed implementation of the scenario using BTs is described in the paper by Iovino et al. (2024). We present the resulting tree and briefly explain its structure.

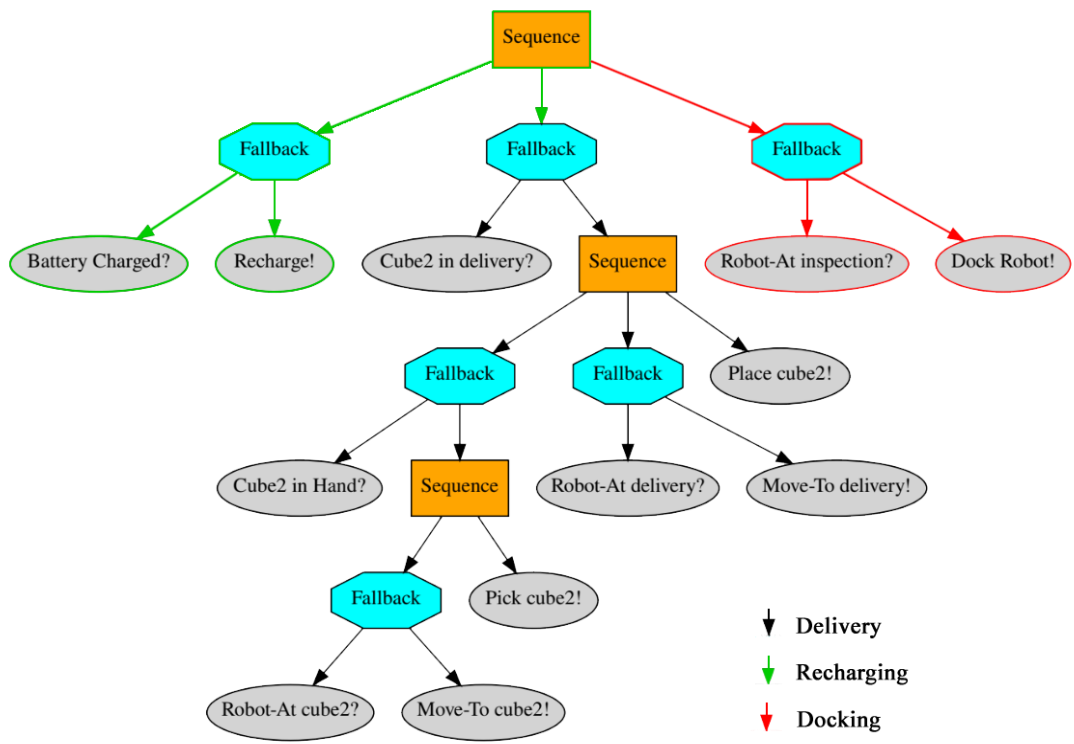

*Figure 1. BT structure for the robot control scenario: black arrows represent the base delivery task, green arrows represent the priority recharging task, and red arrows represent the docking task (source: Iovino et al., 2024).*

### Base Task: Delivery

The base delivery task - moving the object to the target location - is implemented by the central subtree (black arrows). At each tick, the tree is traversed top-down and left-to-right until the "Place cube2!" node is executed.

### Extension 1: Recharging Task

The priority recharging task is implemented by adding a new root node, to which a Fallback node and two leaf nodes are attached on the left: one that checks the battery level and another that performs recharging (green arrows).

The key mechanism at play here is polling-based reactivity. At every tick, the system descends from the root and first evaluates the battery state. If the battery level is sufficient, the left branch returns Failure, and control proceeds to the delivery task. If the battery is low, the left branch executes, preempting any ongoing actions in the right branch.

### Extension 2: Docking Task

The docking task is also implemented by adding a new subtree (red arrows). Since a Sequence root node already exists, modifying the BT involves: (1) creating a Fallback node, (2) adding two leaf nodes to the control node ("Robot-At inspection?" and "Dock Robot!"), and (3) appending the new subtree to the root. Again, the remainder of the tree remains unaffected by this addition.

## 3.3. EO Implementation

We now detail how the scenario is implemented using EO models[2].

### Base Task: Delivery

Robot behavior for delivering an object to a target location is implemented in EO using the Model Delivery:

```
DeliveryTask: Model: Model Delivery
: Relation: robot
: Relation: robotLoc
: Relation: objectLoc
: Relation: targetLoc
```

---

[2] Fully functional models are written in the Boldsea Semantic Language and interpreted (without compilation into code) by the boldsea framework. Model and vocabulary input, editing, and execution are performed by an analyst via the framework's IDE interfaces. Simple user interface models for simulating delivery are provided in the Appendix A. The simulation results are presented in the video: https://youtu.be/XOcELqSfZrI.

```
    : Attribute: cameObjectLocation
    :: Condition: $.robotLoc != $.objectLoc && $.objectLoc != $.robot
    :: SetDo: $.robotLoc <- $.objectLoc
    : Attribute: took
    :: Condition: $.robotLoc == $.objectLoc && $.robotLoc != $.targetLoc
    :: SetDo: $.objectLoc <- $.robot
    : Attribute: cameTargetLocation
    :: Condition: $.robotLoc != $.targetLoc && $.objectLoc == $.robot
    :: SetDo: $.robotLoc <-  $.targetLoc
    : Attribute: put
    :: Condition: $.robotLoc == $.targetLoc  &&  $.objectLoc == $.robot
    :: SetDo: $.objectLoc <- $.targetLoc
    : Attribute: delivered
    :: SetValue: $.objectLoc == $.targetLoc

    # Note: for brevity, the system acts SetDo are presented in a simplified form.
    # The full syntax is provided in Appendix A.
```

Model Operation Description:

The delivery task individual (a specific instance of the process) created according to Model Delivery establishes the binding to the robot (`robot`) and records the current locations of the robot (`robotLoc`), the object (`objectLoc`), and the target (`targetLoc`). All three locations can be arbitrarily modified by an operator during the delivery process.

The attributes `cameObjectLocation`, `took`, `cameTargetLocation`, and `put` are model events that become available to an actor (user or system) when their respective conditions are satisfied. This simulates sensor triggers indicating the successful completion of corresponding actions: the robot reaching the object's location, grasping the object, arriving at the target location, and placing the object.

Upon "sensor activation" (assigning the value "1" to the attribute), the `SetDo` event modifies the relationships `robotLoc`, `objectLoc`, and `targetLoc`, causing the robot to execute the corresponding action: movement, grasping, or placement. Grasping is simulated by setting the object's location to the robot's own location (`$.objectLoc <- $.robot`).

Success is defined as the robot delivering the object to the target location (`$.objectLoc == $.targetLoc`) within several steps, starting from any initial configuration. Success is recorded by setting the `delivered` indicator to "1". Any operator intervention modifying the object's or target's location does not prevent the successful outcome.

Based on this description, the core principle of the dataflow execution mechanism for event-based models can be succinctly expressed: when the condition of one model event is satisfied, it changes the system state (object positions), which automatically triggers the condition of another event, cascading until the planned result is achieved[3].

## Extension 1: Recharging Scenario

The task of switching the robot to recharging could have been addressed by extending the `Model Delivery`, thereby reducing the number of new model events introduced. However, we propose a solution that better aligns with the EO architecture: introducing a new independent action, `RechargingTask (Model Recharging)`, along with an orchestrating model - `Model Robot`.

`Model Robot` contains attributes that monitor battery status: `batteryMin` (minimum threshold), `batteryLevel` (current level), and `batteryLow` (a Boolean indicator). The task attribute acts as a trigger, dynamically assigning either `Model Recharging` or `Model Delivery` based on the value of `batteryLow`.

```
Robot: Model: Model Robot
 : Relation: location
 : Relation: task
 :: SetValue: $.batteryLow ?
    $($EQ.$Model("Model Recharging"), $EQ.robot($CurrentIndividual)) :
    $($EQ.$Model("Model Delivery"), $EQ.robot($CurrentIndividual))
 : Relation: station
 : Attribute: batteryMin
 : Attribute: batteryLevel
 : Attribute: batteryLow
 :: SetValue: +$.batteryLevel  < +$.batteryMin
```

Model Recharging mirrors the structure of Model Delivery:

- It directs the robot to the charging station location and monitors the charging status via the charging attribute.
- The charged attribute serves as an interface indicator for the battery state.
- Upon reaching full charge (`batteryLevel = 100`), `Model Robot` automatically switches the task back to `Model Delivery`, resuming the delivery task from the robot's current location (the station), using the previously defined object and target locations.

```
RechargingTask: Model: Model Recharging
 : Relation: robot
 : Relation: robotLoc
 :: SetValue: $($.robot).location
```

---

[3] A formal justification for the executability of algorithms in the event-based dataflow paradigm is provided in Boldachev (2025, Section 2).

```
    : Relation: targetLoc
    :: SetValue: $($.robot).station
    : Attribute: cameTargetLocation
    :: Condition: $.robotLoc != $.targetLoc
    :: SetDo: $.robot.location = $.targetLoc
    : Attribute: charging
    :: Condition: $.robotLoc == $.targetLoc && $.charged != 100
    :: SetDo: $.robot.batteryLevel <- 100
    : Attribute: charged
    :: SetValue: $($.robot).batteryLevel
```

Switching between tasks is driven by dataflow reactivity - the automatic propagation of state changes through the semantic graph - rather than by periodic polling of system components.

Extension 2: Docking Scenario

The docking task is implemented by adding a new relation (`dock`) to `Model Robot` and appending a `SetDo` restriction to the delivered attribute in `Model Delivery`. Upon successful delivery (`delivered == true`), this triggers an automatic command for the robot to navigate to the dock location.

```
Robot: Model: Model Robot
   …
: Relation: dock

DeliveryTask: Model: Model Delivery
   …
: Attribute: delivered
:: SetValue: $.objectLoc == $.targetLoc
:: SetDo: $.robotLoc <- $($.robot).dock
```

# 4. Comparative Analysis of Implementations Based on Key Metrics

The comparison of the scenario implementations will be conducted using the key metrics proposed in Iovino et al. (2024) - modularity, reactivity, and readability - adapted for comparing the architectural paradigms of Behavior Trees (BTs) and Executable Ontologies (EO). We additionally include an analysis of system component unification, a unique characteristic of EO.

## 4.1. Formal Mapping

In BTs, the units for measuring size and complexity are nodes (composite or leaf) and edges. To enable a valid comparison with EO, appropriate correspondences must be established within the semantic models. The most suitable unit of measurement is the root-level model events (`Attribute, Relation`). Nested restricting properties of model events - such as `Condition`, `SetValue`, and `SetDo` - should be regarded as elements that enable control and data retrieval, not as independent nodes (in BTs, they correspond to the program code within leaf nodes).

Base Task: Delivery

Implementing the base delivery task in BTs requires creating a root node (Fallback) and coordinating several leaf condition nodes (checks) and leaf action nodes (move-to, pick-up, place). The delivery subtree (Fig. 1, black arrows) comprises 14 nodes and 13 edges.

In EO, the task is implemented as the `Model Delivery`, consisting of 8 elements: four controlling model events, three foundational relations (`robotLoc, objectLoc, targetLoc`), and one indicator (`delivered`). The execution logic is explicitly represented as semantic restricting properties (`Condition, SetValue, SetDo`).

A comparison of the base metrics for this task reveals comparable complexity in terms of element count.

| Metric | BTs | EO (Delivery Task) |
| --- | --- | --- |
| Total Nodes | ≈ 7–9 | 7–8 |
| Action Nodes | 3–4 | 4 |
| Coordination Depth | 3–4 levels | 2 levels (event → restrictions) |
| Execution Mechanism | Tree Tick | Condition (dataflow) |

*Table 1. Metric comparison for the base Delivery task.*

The key distinction is qualitative, not quantitative. In BTs, checks and action effects are encapsulated within the hidden code of leaf nodes. In EO, the logic is explicit and intrinsic to the model itself - encoded directly in `Condition`, `SetValue`, and `SetDo` expressions - fundamentally enhancing observability and verifiability.

Extension 1: Recharging Task

Integrating recharging into BTs required creating four new nodes (a root node and the nodes in the left subtree) and attaching the existing delivery subtree to the new root (Fig. 1, green arrows).

In EO, integrating the recharging scenario is achieved by introducing the new Model Recharging alongside the existing Model Delivery, without modifying the latter. The switch between models is orchestrated by extending `Model Robot`: adding the battery monitoring events (`batteryMin, batteryLevel, batteryLow`) and an orchestrating child event, `SetValue`, to the task relation. In total, four model action events and six condition events were added.

However, such a count is misleading, as the fundamental nature of a node in BTs (hidden logic within code) differs qualitatively from that in EO (explicit logic in `Condition/SetValue`).

### Extension 2: Docking Task

Adding docking to the BT after delivery requires creating a new subtree with three nodes (Fig. 1, red arrows). In EO, this task is solved by adding one `SetDo` restriction to `Model Delivery` and introducing a new relation (dock) to specify the docking station location.

It should be acknowledged that comparing technologies based on the number of structural elements (nodes and edges) - a cornerstone metric in the BT vs. FSM comparison (Iovino et al., 2024) - reveals no fundamental difference between BTs and EO. This is true unless one considers that the EO model not only formalizes the control structure but also executes it without requiring additional procedural code.

## 4.2. Modularity: Structural Composition vs. Semantic Orchestration

Modularity assesses the ease of adding, removing, or modifying functional components without disrupting the rest of the logic. Consider the process of adding the priority recharging task to the existing delivery task.

In BTs, adding new functionality requires modifying the root structure: a new root node is created, the recharging subtree is inserted as the left (highest-priority) branch, and the existing delivery tree becomes the right branch. According to the analysis in (Iovino et al., 2024), the complexity of this operation is $O(1)$, an excellent result. This exemplifies *structural modularity*, based on the composition of procedural blocks.

In EO, this process manifests as *semantic orchestration*. Model Recharging is created as a completely independent, standalone semantic module, structurally unrelated to Model Deliver. The sole modification is adding a condition to Model Robot that activates different tasks. This exemplifies *semantic modularity*, where independent models are coordinated through changes to declarative rules. The effort required for modification is comparable to, or even less than, that in BTs, as it avoids restructuring the control-flow hierarchy.

**Conclusion:** Both technologies demonstrate high modularity. However, BTs achieve *structural modularity* through the *composition of procedures* within a hierarchical structure, whereas EO achieves *semantic modularity* through the *orchestration of independent models* via declarative rules. The EO approach potentially offers greater flexibility, as the orchestration logic is entirely decoupled from the models of specific tasks.

## 4.3. Reactivity: Tick-Based Polling versus Event-Driven Dataflow Execution

Reactivity refers to a system's ability to interrupt ongoing actions to execute higher-priority tasks. In our scenario, this means interrupting delivery to initiate recharging.

In BTs, reactivity is achieved through periodic, tick-based evaluation. At every tick, the tree is traversed anew from the root. The battery state check, residing in the left (highest-priority) branch, is always evaluated before any delivery-related actions in the right branch. If the low-battery condition becomes true, execution immediately switches to the recharging subtree, preempting any currently running lower-priority task.

In EO, reactivity is enabled by event-driven dataflow, which operates without polling through subscriptions triggered by `Condition` and `SetValue` expressions. The EO engine reacts automatically to changes in the `batteryLevel` value, which in turn triggers the `SetValue` rule in `Model Robot` to switch the `task` to `Model Recharging`.

**Conclusion:** Both approaches achieve full reactivity. However, BTs rely on a procedural mechanism of continuous state polling, while EO employs a declarative, event-driven dataflow mechanism. The EO approach is more efficient in principle, since computations occur only in response to relevant data changes, not at every fixed tick.

### 4.4. Readability and Design Principles: Algorithm versus Model

Readability measures how easily a human can understand a system's behavior by examining its visual representation.

The primary advantage of BTs lies in the high readability of their control flow, as their intuitive graphical tree structure is easily interpreted as a sequential algorithm.

While BTs offer superior readability for the *control flow*, EO provides superior readability for the *domain model* and its underlying rules. The logic is not hidden within code blocks but is explicitly declared as conditions and constraints, making the rules of the world transparent and directly inspectable. Behavior emerges from the cascading execution of declarative rules (`SetValue`, `Condition`) triggered by data changes, rather than from a sequential traversal of command nodes.

**Conclusion:** BTs excel in readability of the *process*, while EO offers superior readability of the *domain rules*. The choice depends on whether transparency of the algorithm or accuracy of the world model is more critical for the task. Furthermore, EO's declarative nature provides a key design advantage: it is easier to define the *conditions under which an event is possible* (dataflow) than to precompute and enumerate all its potential consequences (control flow).

### 4.5. Bridging the Semantic-Process Gap: Unification of Data, Logic, and UI

An additional distinction between EO and BTs - beyond standard comparison metrics - directly addresses the semantic-process gap.

In BTs, the control structure (the tree) is separate from the state representation (often a blackboard or external database), the behavior logic (implemented in Python/Node.js code), and the user interface (a separate UI framework). EO unifies all these aspects: the semantic graph *is* the state, the restrictions *are* the logic, and the UI can be auto-generated from the same models. This eliminates integration points and reduces development complexity from a multi-layer system to a single declarative model.

**Conclusion:** This unified approach dramatically reduces the complexity of system development and integration. All aspects - from data structure to interface buttons - are described and governed within a single, event-based semantic model.

## 4.6. Comparative Analysis of Foundational Principles

The analysis demonstrates that Behavior Trees (BTs) and Executable Ontologies (EO) are not merely different technologies, but represent fundamentally different approaches to design. The table below summarizes their core architectural differences.

| Aspect | BT | EO |
| --- | --- | --- |
| Coordination | Hierarchical tree with Sequence/Fallback nodes; logic encapsulated within leaf node code | Declarative model events governed by restricting properties (`Condition`, `SetValue`, `SetDo`) |
| Execution | Centralized tick mechanism performing top-down tree traversal | Distributed activation via event-driven dataflow subscriptions triggered by state changes |
| System State | Partially stored in a blackboard or external systems; current position = active tree node | State represented as a temporal graph of individuals and their properties; behavior emerges from computation over the current graph state |
| Temporality | External logging (not inherent to the architecture) | Native temporal graph with full history: actor, cause, and timestamp for every event |
| Interface | Separate UI development; independent of the behavior model | UI auto-generated from semantic models; UI elements are part of the ontology |
| Interoperability | Integration at the code and API level | Semantic interoperability via shared vocabularies; model composition without code modification |

*Table 2. Architectural mechanisms*

Table 2 clearly illustrates that the choice between BTs and EO is not merely a technical decision, but a conceptual one that defines the fundamental approach to developing robotic systems.

## 4.7. Operational Properties

Beyond their fundamental architectural differences, BTs and EO exhibit significant distinctions in operational characteristics critical for production systems.

The ability to modify the system without downtime (zero downtime updates) is a key advantage of EO. Since the system state is stored as facts in the event graph - not as a "current position" - adding or modifying models does not interrupt ongoing processes. In contrast, BTs typically require recompilation and restart, risking loss of execution context.

In EO, observability and reproducibility follow directly from the dataflow engine operating over a temporal event graph: every robot act is executed only when all conditioning events, including sensor readings, are already present in the graph. From the engine's perspective, observable behavior is nothing more than a particular traversal of this recorded event history. In BTs, observability is limited to logging node execution without deep semantic context.

Table 3 summarizes the key operational differences between the two approaches.

| Property | BT | EO |
| --- | --- | --- |
| Modification without Downtime | Requires system recompilation and restart; risk of losing current execution state | Runtime model modification without interruption; state preserved in the event graph |
| Observability | Logging of node execution; limited semantic context | Full semantic traceability: each event is linked to its model, actor, and causal predecessors |
| Reproducibility | Partial: behavior depends on internal state and ad-hoc logs; exact replay is rarely guaranteed | Complete: replay from the temporal event graph with unambiguous reconstruction of engine state and robot behavior |
| Effort for Extensions | Adding a subtree requires restructuring; a new node = new code | Adding an independent model; often requires modifying only one condition (e.g., Robot.task) |
| Validation | Structural correctness of the tree (static) | Runtime validation of each event against its model: data types, constraints, access permissions |

*Table 3. Operational properties*

In EO this reproducibility refers to the internal decision process: given the same sequence of external events, the engine can reconstruct which rules fired and which values were produced at each step. It does not remove nondeterminism originating from sensors or the physical environment.

It is important to note that some entries in Tables 2 and 3 reflect common patterns in current BT frameworks and in our specific EO implementation, rather than hard theoretical limits of the approaches. For example, zero downtime updates and full temporal traceability are architectural goals that EO supports natively, whereas BT frameworks could in principle be extended with similar capabilities. Our claim is therefore not that BTs cannot realize these properties at all, but that EO makes them first-class, semantically grounded features rather than add-on mechanisms.

## 5. Discussion

### 5.1. Interpretation of results

Our analysis reveals a critical methodological limitation: metrics designed to compare architectures *within* the control-flow paradigm (e.g., BTs vs FSM) are insufficient for a full comparison of the paradigms themselves (*control-flow* vs *dataflow*). Structural metrics (node count) are misleading because they equate a code block (BT leaf) with a semantic constraint (EO restriction). The true difference lies in the *level of abstraction*: BTs model an *algorithm*, while EO models a *semantic domain*. This explains why adding a new task requires modifying the *orchestration rule* (one point) in EO, not restructuring the *control tree* (multiple points) in BTs. The "cost" is not in the number of elements, but in the *nature of the change* - changing rules versus changing the control flow.

These limitations are not a flaw of the original metrics, which were designed for intra-paradigm comparisons such as BTs versus FSMs, but they do indicate that contrasting control-flow and dataflow architectures requires a distinct methodological framework. Key metrics may include design costs, which reflect the complexity of defining activation conditions in the EO approach compared to the effort required to construct a complete algorithm in BT. Operational characteristics are also important, encompassing computational load, where the BT tick mechanism is compared with the event-driven structure of EO and its associated overhead. Scalability may become a critical metric, characterizing the behavior of both architectures as the number of parallel tasks increases and the temporal graph volume grows.

This study is intentionally qualitative and architectural in nature. We do not present systematic performance benchmarks, robustness experiments, or user studies; instead, we rely on a single, well-specified scenario as a comparative case study. Quantitative evaluation of BT and EO implementations - e.g., in terms of execution time, memory footprint, and development effort - remains an important direction for future work.

### 5.2. Characteristics of Executable Ontologies

The comparative analysis has revealed the key architectural characteristics of EO:

- **Temporal Transparency:** The EO architecture automatically generates a complete temporal graph of all recorded events, which is critical for auditing, debugging, and scenario reproduction.

- **Model-Based Validation:** The model-based principle ensures events are created only according to predefined models, guaranteeing continuous semantic integrity and real-time data validation by the engine.
- **Unified Language:** EO provides a single formalism for describing data, logic, and even user interfaces.
- **Semantic Interoperability:** By nature, EO promotes solution reuse through shared vocabularies, enabling seamless interaction between independent models.
- **Natural Parallelism:** EO models, being based on the dataflow architecture, inherently support the parallel execution of tasks.
- **Zero downtime Modularity:** EO allows the addition of new functionality without stopping or rewriting existing code.

## 5.3. Limitations and Challenges

Despite these advantages, EO's declarative approach presents challenges:

**Cognitive barrier:** EO introduces a new way to model IT systems. Developers face cognitive challenges when transitioning from imperative programming (think "steps") to declarative modeling (think "rules of the world"), requiring training and a change in design methodology.

**Immature technology and performance concerns:** Comprehensive data on EO engine testing under industrial (production) conditions is currently lacking. Questions regarding reliability, as well as performance and scalability when handling large temporal graphs and high computational loads, remain open and require further research.

**Lack of ecosystem:** As a new technology, EO lacks diverse supporting tools (beyond the provided boldsea IDE), libraries of standard models, comprehensive publicly available documentation, and, critically, specialists proficient in the technology. This significantly raises the entry barrier for new teams.

## 5.4. Recommendations for Application

The choice between BTs and EO should be dictated by the nature of the task. Behavior Trees remain the ideal choice for systems with well-defined, predictable algorithms. If a robot's task involves executing a complex but rigid sequence of actions in a relatively controlled environment (e.g., a single robot on a production line), BTs offer superior process readability and a mature ecosystem.

EO is suited for dynamic, adaptive, and evolving systems. It is appropriate for multi-agent scenarios (where multiple robots, humans, and sensors asynchronously generate events using the temporal graph as a shared data bus), for systems requiring semantic interoperability (integration with diverse IT systems via shared models and vocabularies), and where zero downtime updates are a critical business requirement.

Hybrid systems represent a promising direction, where BTs handle execution-level tasks by efficiently implementing rigidly programmed "skills," while EO acts as the high-level semantic orchestrator and domain model. This approach represents a promising direction for future research.

Another direction is using executable ontologies in semantic robotics (Bernardo et al., 2025, Zhao et al., 2024). In this area it is possible to generalize the temporal semantic graph to a multi-subject setting, where different human, robotic, and software actors perceive and manipulate the shared event history through their own semantic models, effectively using the graph as a communication bus. Using executable ontologies in semantic robotics warrants separate analysis

# 6. Conclusion

This paper demonstrates that Behavior Trees and Executable Ontologies are not competing technologies, but represent complementary architectures for robotic control. BTs remain an excellent tool for implementing precise, predictable, step-by-step algorithms. EO, however, proposes a change in approach: it models the world in which the robot operates, not the algorithm it follows. This shift enables unique operational advantages: non-invasive runtime extensibility, full semantic traceability, and the unification of data, logic, and interface within a single, verifiable model. The choice between BTs and EO is not merely technical, but philosophical: it is a choice between an algorithm that dictates behavior, and a world model from which behavior emerges. Future work will focus on integrating these architectures in hybrid systems and leveraging Large Language Models (LLMs) for the automatic generation and modification of EO models from natural language specifications.

The key insight from this research is the understanding that comparing paradigmatically distinct architectures requires the development of a new methodology. Metrics used for comparing BTs and FSMs proved insufficient for a full evaluation of the differences between control-flow and dataflow approaches. Further research is needed into operational characteristics (performance, scalability), economic indicators (development and maintenance costs), and a systematic analysis of applicability across different classes of robotics tasks.

# References

1. Boldachev, A. (2025). Executable Ontologies: Synthesizing Event Semantics with Dataflow Architecture. arXiv:2509.09775 [cs.AI]. Retrieved from https://arxiv.org/abs/2509.09775
2. Boldachev, A. (2025). Subject-Event Ontology Without Global Time: Foundations and Execution Semantics. arXiv:2510.18040 [cs.AI]. Retrieved from https://www.arxiv.org/abs/2510.18040
3. Iovino, M., Förster, J., Falco, P., Chung, J. J., Siegwart, R., and Smith, C. (2024 ). Comparison between Behavior Trees and Finite State Machines. arXiv:2405.16137v1 [cs.RO] Retrieved from https://arxiv.org/pdf/2405.16137
4. Bryan, S. (2024 ). Executable ontologies: How to empower expert knowledge workers with AI language models. Shepbryan.com. Retrieved from

https://www.shepbryan.com/blog/executable-ontologies-empower-expert-knowledge-workers-with-ai-language-models

5. Colledanchise, M. and Ögren, P. (2018 ). Behavior Trees in Robotics and AI: An Introduction. CRC Press, Jul. 2018, kTH. [Online]. Available: https://www.taylorfrancis.com/books/9780429950902
6. Fowler, M. (2005 ). Event sourcing. Martinfowler.com. Retrieved from https://martinfowler.com/eaaDev/EventSourcing.html
7. Gugliermo, S., Domínguez, D. C., Iannotta, M., Stoyanov, T., and Schaffernicht, E. (2024 ). Evaluating behavior trees. Robotics and Autonomous Systems, p. 104714. [Online]. Available: https://www.sciencedirect.com/science/article/pii/S0921889024000976
8. Hasselbring, W., Wojcieszak, M., and Dustdar, S. (2021 ). Control Flow Versus Data Flow in Distributed Systems Integration: Revival of Flow-Based Programming for the Industrial Internet of Things. IEEE Internet Comput. 2021, 25, 5–12.
9. Colledanchise, M. and Ögren, P. (2016). How Behavior Trees Modularize Hybrid Control Systems and Generalize Sequential Behavior Compositions, the Subsumption Architecture, and Decision Trees. IEEE Transactions on Robotics, 32(6), 1483–1499.
10. Hong, W., Liu, Y., and Zhang, X. (2023). Formal Verification based Synthesis for Behavior Trees. In 2023 IEEE 34th International Symposium on Software Reliability Engineering (ISSRE).
11. Bernardo, R., Sousa, J. M. C., and Gonçalves, P. J. S. (2025). Ontological framework for high-level task replanning for autonomous robotic systems. Robotics and Autonomous Systems, 183, 104870.
12. Aguado, E., Gomez, V., Hernando, M., Rossi, C., and Valero, M. (2024). A survey of ontology-enabled processes for dependable robot autonomy. Frontiers in Robotics and AI, 11.
13. Zhao, J., Zhang, Z., and Li, Y. (2024). Ontology Based AI Planning and Scheduling for Robotic Systems. In 2024 IEEE International Conference on Robotics and Automation (ICRA).
14. Alenzi, Z.; Alenzi, E.;Alqasir, M.; Alruwaili, M.; Alhmiedat,T.; Alia, O.M. A Semantic Classification Approach for Indoor Robot Navigation. Electronics 2022,11, 2063. Appendices

# Appendices

## Appendix A.  Scenarios Code

We present the comprehensive EO code that implements the Delivery, Recharging, and Docking Scenarios. The code, as presented, is loaded into the graph via the engine's console. Upon loading, it undergoes full validation for both format and semantic consistency against the previously loaded code (ensuring the presence of required concepts, properties, model events, and individuals).

For brevity and clarity, the code does not include the `Permission` restricting properties, which manage access rights for the robot, administrator, or sensors. Furthermore, the demonstration is limited to one robot and one object and does not contain events for creating new individuals. The code for `SetDo` system acts is provided in full, rather than as pseudocode, as seen in the examples in the main text.

User Interface (UI) pages are described in BSL as individuals of the `View` concept models and are interpreted by the View-controller. The properties of the View-models are not listed, as they are part of the genesis block of system events in the graph, which is loaded when the engine starts.

The code is divided into three blocks corresponding to the scenarios. Each subsequent block is loaded into the graph without stopping the execution of the previous scenario (i.e., "hot-loaded") with minimal write latency. Repeated initializations of models and individuals are not performed; instead, the code references existing elements to which new events are to be attributed.

```
# DELIVERY SCENARIO
# CONCEPTS
Concept: Instance: Robot
Concept: Instance: Location
Concept: Instance: DeliveryTask
# PROPERTIES
Relation: Individual: robot
: Range: Robot
Relation: Individual: robotLoc
: Range: Location
Relation: Individual: targetLoc
: Range: Location
Relation: Individual: objectLoc
: Range: Location
Relation: Individual: location
: Range: Location
Relation: Individual: task
Attribute: Individual: delivered
: DataType: Boolean
Attribute: Individual: put
: DataType: Boolean
Attribute: Individual: cameTargetLocation
: DataType: Boolean
Attribute: Individual: took
: DataType: Boolean
Attribute: Individual: cameObjectLocation
: DataType: Boolean

# MODELS
DeliveryTask: Model: Model Delivery
: Relation: robot
:: Immutable: 1
: Relation: robotLoc
:: SetValue: $($.robot).location
: Relation: objectLoc
: Relation: targetLoc
: Attribute: cameObjectLocation
:: Condition: $.robotLoc != $.objectLoc
   && $.objectLoc != $.robot
:: SetDo: ({'$do':'EditIndividual',
   '$IndividualID': $.robot,
   '$Condition': $Value === "1",
   'location': $.objectLoc})
: Attribute: took
:: Condition: $.robotLoc == $.objectLoc
   && $.robotLoc != $.targetLoc
:: SetDo: ({ '$do':'EditIndividual',
   '$IndividualID': $CurrentIndividual,
   '$Condition': $Value == "1",
   'objectLoc': $.robot})
: Attribute: cameTargetLocation
:: Condition: $.robotLoc !=
   $.targetLoc && $.objectLoc == $.robot
:: SetDo: ({'$do':'EditIndividual',
   '$IndividualID': $.robot,
   '$Condition': $Value === "1",
   'location': $.targetLoc})
: Attribute: put
:: Condition: $.robotLoc == $.targetLoc
   &&  $.objectLoc == $.robot
:: SetDo: ({ '$do':'EditIndividual',
   '$IndividualID': $CurrentIndividual,
   '$Condition': $Value == "1",
   'objectLoc': $.targetLoc})
```

```
: Attribute: delivered
:: SetValue: $.objectLoc == $.targetLoc

Robot: Model: Model Robot
: Relation: location
: Relation: task
:: SetValue: $($EQ.$Model("Model
   Delivery"),
   $EQ.robot($CurrentIndividual))

Location: Model: Model Location

# INDIVIDUALS
Location: Individual: Loc A
: SetModel: Model Location
Location: Individual: Loc B
: SetModel: Model Location
Location: Individual: Loc C
: SetModel: Model Location

Robot: Individual: Robot 1
: SetModel: Model Robot
: location: Loc A

DeliveryTask: Individual: Delivery 1
: SetModel: Model Delivery
: robot: Robot 1

# VIEW
View: Model: Model View Individual
: Attribute: ConceptPage
: Relation: IndividualID
: Attribute: ViewConcept
:: Relation: IndividualList
::: SetValue: $.IndividualID
:: Attribute: ViewMode
:: Attribute: Title
:: Attribute: Include
:: Attribute: Exclude
:: Attribute: Control
::: Attribute: Title
::: Attribute: ControlType
::: Attribute: Value

View: Individual: View Delivery
: SetModel: Model View Individual
: ConceptPage: DeliveryTask
: IndividualID: Delivery 1
: ViewConcept: DeliveryTask
:: IndividualList: Delivery 1
:: ViewMode: showcase
```

```
:: Control: cameObjectLocation
::: Title: Came Object Location
::: ControlType: button
::: Value: 1
:: Control: took
::: Title: Took Object
::: ControlType: button
::: Value: 1
:: Control: cameTargetLocation
::: Title: Came Target Location
::: ControlType: button
::: Value: 1
:: Control: put
::: Title: Put Object
::: ControlType: button
::: Value: 1

View: Individual: View Robot
: SetModel: Model View Individual
: ConceptPage: Robot
: IndividualID: Robot 1
: ViewConcept: Robot
:: IndividualList: Robot 1
:: ViewMode: showcase

# RECHARGING SCENARIO
# CONCEPT
Concept: Instance: RechargingTask

# PROPERTIES
Relation: Individual: station
: Range: Location
Attribute: Individual: batteryLevel
: DataType: Number
Attribute: Individual: batteryMin
: DataType: Number
Attribute: Individual: batteryLow
: DataType: Boolean
Attribute: Individual: charging
: DataType: Boolean
Attribute: Individual: charged
: DataType: Number

# MODELS
Robot: Model: Model Robot
: Relation: task
:: SetValue: $.batteryLow ?
   $($EQ.$Model("Model Recharging"),
   $EQ.robot($CurrentIndividual)) :
```

```
    $($EQ.$Model("Model Delivery"),
    $EQ.robot($CurrentIndividual))
: Relation: station
:: Immutable: 1
: Attribute: batteryMin
:: ValueCondition: $Value > 0 &&
    $Value < 100
: Attribute: batteryLevel
:: ValueCondition: $Value >= 0 &&
    $Value <= 100
: Attribute: batteryLow
:: SetValue: +$.batteryLevel  <
  +$.batteryMin

# Task Recharging
RechargingTask: Model: Model Recharging
: Relation: robot
:: Immutable: 1
: Relation: robotLoc
:: SetValue: $($.robot).location
: Relation: targetLoc
:: SetValue: $($.robot).station
: Attribute: cameTargetLocation
:: Condition: $.robotLoc != $.targetLoc
:: SetDo: ({ '$do':'EditIndividual',
  '$IndividualID': $.robot,
  '$Condition': $Value == "1",
  'location': $.targetLoc})
: Attribute: charging
:: Condition: $.robotLoc == $.targetLoc
    && $.charged != 100
:: SetDo: ({'$do':'EditIndividual',
  '$IndividualID': $.robot,
  '$Condition': $Value === "1",
  'batteryLevel': 100})
: Attribute: charged
:: SetValue: $($.robot).batteryLevel

# INDIVIDUALS
Location: Individual: Loc Station
: SetModel: Model Location

Robot: Individual: Robot 1
: SetModel: Model Robot
: station: Loc Station
: batteryMin: 20
: batteryLevel: 100

RechargingTask: Individual: Recharging
: SetModel: Model Recharging
: robot: Robot 1
```

```
# VIEW
View: Individual: View Recharging
: SetModel: Model View Individual
: ConceptPage: RechargingTask
: IndividualID: Recharging
: ViewConcept: RechargingTask
:: IndividualList: Recharging
:: ViewMode: showcase
:: Control: cameTargetLocation
::: Title: Came Target Location
::: ControlType: button
::: Value: 1
:: Control: charging
::: Title: Recharging
::: ControlType: button
::: Value: 1
```

```
# DOCKING SCENARIO

# PROPERTIES
Relation: Individual: dockLoc
: Range: Location
Relation: Individual: dock
: Range: Location

# MODELS
Robot: Model: Model Robot
: SetModel: Model Robot
: Relation: dock
:: Immutable: 1

DeliveryTask: Model: Model Delivery
: SetModel: Model View Individual
: Attribute: delivered
:: SetValue: $.objectLoc == $.targetLoc
:: SetDo: ({'$do':'EditIndividual',
  '$IndividualID': $.robot, '$Condition':
    $Value === "1", 'location':
    $($.robot).dock})

# INDIVIDUALS
Location: Individual: Loc Dock
: SetModel: Model Location

Robot: Individual: Robot 1
: SetModel: Model Robot
: dock: Loc Dock
```

# Appendix B. UI Screenshots[4]

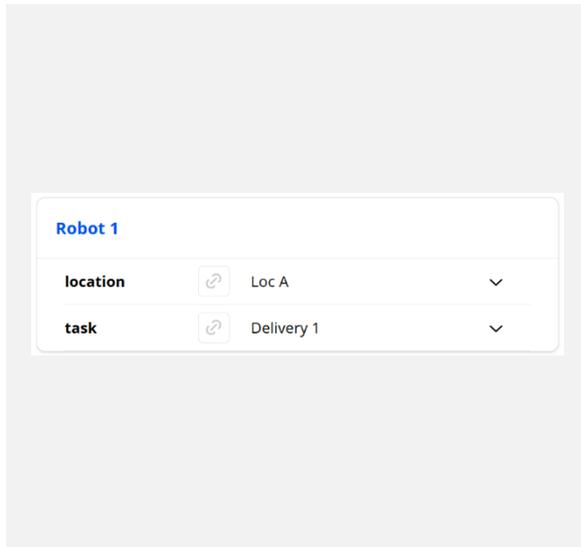

*Figure 2. Robot UI, Delivery Scenario.*

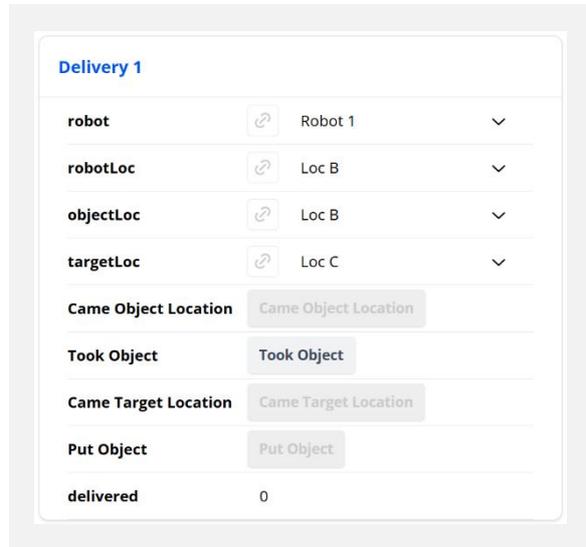

*Figure 3. Delivery Scenario UI (robot at the object's location, awaiting confirmation of its capture).*

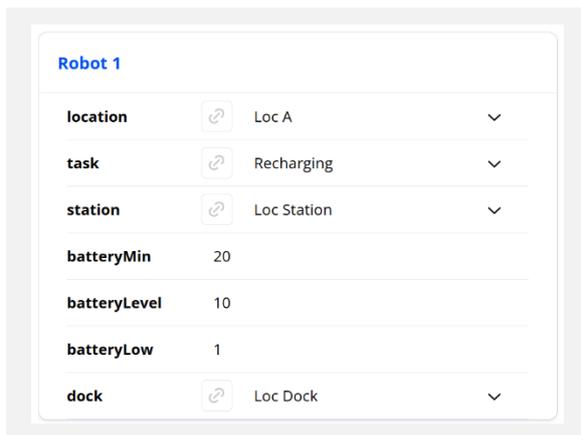

*Figure 4. Robot UI, Recharging Scenario (the battery charge level is below critical; the task has switched to Recharging).*

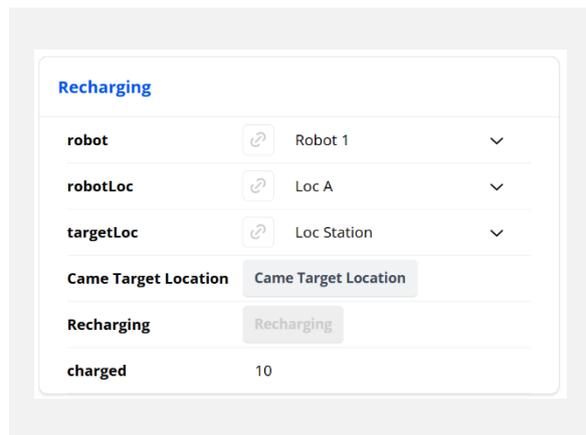

*Figure 5. Recharging Scenario UI (the robot is heading towards the station).*

---

[4] The simulation results are presented in the video: https://youtu.be/XOcELqSfZrI.

*Figure 6. boldsea IDE, Temporal Graph.*

*Figure 7. boldsea IDE, Model Editor.*

*Figure 8. boldsea IDE, Restriction Editor.*